# Multi-Scored Sleep Databases: How to Exploit the Multiple-Labels in Automated Sleep Scoring

Luigi Fiorillo[1,2,*,†], Davide Pedroncelli[3,†], Valentina Agostini[3], Paolo Favaro[1] and Francesca Dalia Faraci[2]

[1]Institute of Informatics, University of Bern, Bern, Switzerland, [2]Institute of Digital Technologies for Personalized Healthcare (MeDiTech), Department of Innovative Technologies, University of Applied Sciences and Arts of Southern Switzerland, Lugano, Switzerland, [3]Department of Electronics and Telecommunications, Politecnico di Torino, Torino, Italy.

Institution where work was performed: Institute of Digital Technologies for Personalized Healthcare (MeDiTech), Department of Innovative Technologies, University of Applied Sciences and Arts of Southern Switzerland, Lugano, Switzerland.

†These authors contributed equally to this work.

*Corresponding author. Luigi Fiorillo, Institute of Digital Technologies for Personalized Healthcare (MeDiTech), Department of Innovative Technologies, University of Applied Sciences and Arts of Southern Switzerland, Lugano, Switzerland. Email: luigi.fiorillo@supsi.ch.

# Abstract

**Study Objectives:** Inter-scorer variability in scoring polysomnograms is a well-known problem. Most of the existing automated sleep scoring systems are trained using labels annotated by a single scorer, whose subjective evaluation is transferred to the model. When annotations from two or more scorers are available, the scoring models are usually trained on the scorer consensus. The averaged scorer's subjectivity is transferred into the model, losing information about the internal variability among different scorers. In this study, we aim to insert the multiple-knowledge of the different physicians into the training procedure. The goal is to optimize a model training, exploiting the full information that can be extracted from the consensus of a group of scorers.

**Methods:** We train two lightweight deep learning based models on three different multi-scored databases. We exploit the label smoothing technique together with a *soft-consensus* ($LS_{SC}$) distribution to insert the multiple-knowledge in the training procedure of the model. We introduce the averaged cosine similarity metric ($ACS$) to quantify the similarity between the hypnodensity-graph generated by the models with-$LS_{SC}$ and the hypnodensity-graph generated by the scorer consensus.

**Results:** The performance of the models improves on all the databases when we train the models with our $LS_{SC}$. We found an increase in $ACS$ (up to 6.4%) between the hypnodensity-graph generated by the models trained with-$LS_{SC}$ and the hypnodensity-graph generated by the consensus.

**Conclusion:** Our approach definitely enables a model to better adapt to the consensus of the group of scorers. Future work will focus on further investigations on different scoring architectures and hopefully large-scale-heterogeneous multi-scored datasets.

Graphical abstract

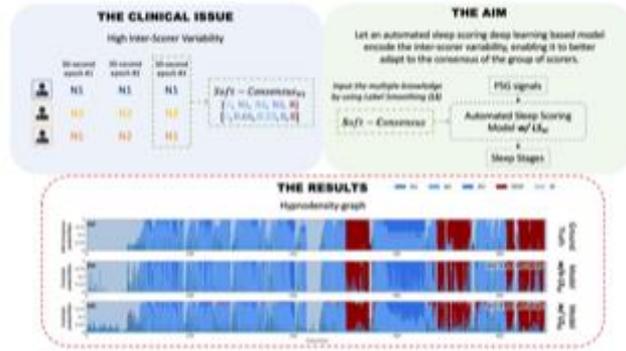

## Statement of Significance

Visual scoring of polysomnography is a highly subjective procedure. Several studies consistently reported the poor agreement between different physicians scoring the same whole-night recording. Existing sleep scoring algorithms, trained on multi-scored databases, overlook to encode in their models the variability among the scorers. We propose a technique to wholly insert the multiple-knowledge of the different physicians into the training procedure of a scoring algorithm. Our approach enables the model to better adapt to the consensus of the group of scorers. Whenever multi-scored databases are available, future researchers should train their models considering the annotations of all the physicians at the same time, rather than averaging their labels and training their algorithm on the averaged consensus.

## Introduction

Sleep disorders represent a significant public health problem that affects millions of people worldwide [1]. Since the late 1950s, the polysomnography (PSG) exam has been the gold standard to study sleep and to identify sleep disorders. It monitors electrophysiological signals such as electroencephalogram (EEG), electrooculogram (EOG), electromyogram (EMG) and electrocardiogram (ECG). The physicians visually extract sleep cycle information from these signals. The whole-night recording is divided in 30-second epochs, and each epoch is classified into one of the five sleep stages (i.e., wakefulness W, stage N1, stage N2, stage N3, and stage REM) according to the AASM guidelines [2]. Worst case scenario, an eight-hour PSG may require up to two hours of tedious repetitive and time-consuming work to be scored. In addition, this manual procedure is highly affected by a low inter-rater scoring agreement (i.e., the agreement between different physicians scoring the same whole-night recording). The inter-rater scoring agreement value ranges from 70% up to slightly more than 80% [3-5]. In [3] the averaged inter-rater agreement of about 83% results from a study conducted on the AASM Inter-scorer reliability dataset, by using sleep stages annotated from more than 2,500 sleep scorers. The agreement was higher than 84% for awake, N2 and REM stages, but it dropped to 63% and 67% for N1 and N3 stages respectively. In fact, the inter-rater agreement varies among sleep stages, patients, sleep disorders and across sleep centers [3], [6].

Since 1960 many different approaches and algorithms have been proposed to automate this time-consuming scoring procedure. Mainly, two different approaches emerged: sleep scoring algorithms learning from well defined features extracted from the knowledge of the experts (shallow learning), and sleep scoring algorithms learning directly from the raw data (deep learning). Thorough reviews about feature based [7-8] and deep learning based [9-10] sleep scoring algorithms can be found in literature. Although the latter algorithms emerged only five years ago, their impressive results have never been reached with the previous conventional feature based approaches. Autoencoders [11], deep neural networks [12], convolutional neural networks [13-20], recurrent neural networks [21-23] and different combinations of them [24-30] have been all proposed only in these last five years.

Almost all of the above algorithms have been trained on recordings scored by a single expert physician. The first remarkable exception comes from [27], where they consider recordings scored by six different physicians [31]. The scoring algorithm was trained on the six-scorer consensus (i.e., based on the majority vote weighted by the degree of consensus from each physician). In [23] the *Dreem* group introduced two publicly-available datasets scored by five sleep physicians. Similarly, they used the scorer consensus to train their automated scoring system. It has been shown that the performance of an automated sleep scoring system is on-par with the scorer consensus [23,27], and mainly that their best scoring algorithm is better than the best human scorer - i.e., the scorer with the higher consensus among all the physicians in the group. Although they both considered the knowledge from the multiple scorers - by averaging their labels and by training their algorithm on the averaged consensus - they still trained the algorithm on a single one-hot encoded label. Indirectly, they are still transferring the best scorer's subjectivity into the model, and they are not explicitly training the model to adapt to the consensus of the group of scorers.

In this work, we train two existing lightweight deep learning-based sleep staging algorithms, our DeepSleepNet-Lite (DSN-L) [32] and SimpleSleepNet (SSN) [23], on three open-access multi-scored sleep datasets. First, we assess the performance of both scoring algorithms trained with the labels given by scorer consensus (i.e., majority vote among the different scorers) and compare it to the performance of the individual scorer-experts. Then we propose to exploit label smoothing along with the *soft-consensus* distribution (*base+LS$_{SC}$*) to insert the multiple-knowledge into the training procedure of the models and to better calibrate the scoring architectures. For the first time in sleep scoring, we are considering the multiple-labels in the training procedure, the annotations of all the scorers are taken into account at the same time. We finally assess the performance and we quantify the similarity between the hypnodensity-graph generated by the models - trained with and without label smoothing - and the hypnodensity-graph generated by the scorer consensus.

In the present work we investigate a different approach in exploiting multi-scored database information. In particular: (1) we demonstrate the efficiency of label smoothing along with the *soft-consensus* distribution in both calibrating and enhancing the performance of both DSN-L and SSN; (2) we show how the model can better resemble the scorer group consensus, leading to a similarity increase between the hypnodensity-graph generated by the model and the hypnodensity-graph generated by the scorer consensus.

## Methods

In this section we first present the three publicly available databases used in this study: IS-RC (Inter-scorer Reliability Cohort) [31]; DOD-H (Dreem Open Dataset - Healthy) and DOD-O (Dreem Open Dataset - Obstructive) [23]. We then briefly describe the architectures of the two deep learning-based scoring algorithms DSN-L [32] and SSN [23]. Next, we show how to compute the consensus in a multi-scored dataset, i.e., how to compute the label among multiple-scorers so as to train our *baseline* algorithms and to be able to evaluate their performance. In **Label smoothing with soft-consensus** subsection we describe in detail how to compute the *soft-consensus* distribution, and how to exploit it along with the label smoothing technique during the training procedure. The aim is to show how to insert the multiple-labels of the different scorers into the training procedure of our algorithms. We finally report all the experiments conducted on both DSN-L and SSN algorithms, i.e., *base, base+LS$_U$ and base+LS$_{SC}$* models, and the metrics exploited to evaluate their performance.

## Datasets

**IS-RC.** The dataset contains 70 recordings (0 males and 70 females) from patients with sleep-disordered breathing aged from 40 to 57. The recordings were collected at the University of Pennsylvania. Each recording includes the EEG derivations C3-M2, C4-M1, O1-M2, O2-M1, one EMG channel, left/right EOG channels, one ECG channel, nasal airway pressure, oronasal thermistor, body position, oxygen saturation and abdominal excursion. The recordings are sampled at 128 Hz.

We only consider the single-channel EEG C4-M1 to train our DSN-L architecture, and we use multi-channel EEG, EOG, EMG and ECG to train the SSN architecture. A band-pass Chebyshev IIR filter is applied between [0.3, 35] Hz. Each recording is scored by six clinicians from five different sleep centers (i.e., University of Pennsylvania, University of Wisconsin at Madison, St. Luke's Hospital (Chesterfield), Stanford University and Harvard University) according to the AASM rules [2]. The dataset contains the following annotations $W$, $N1$, $N2$, $N2$, $R$, and $NC$, where $NC$ is a not classified epoch. Some epochs are not scored by all the six physicians, and even for some of them we don't have any annotation (i.e $NC$). We decided to remove the epochs classified by all the scorers as $NC$. Epochs with less than six annotations are equally taken into account to avoid excessive data loss.

**DOD-H.** The dataset contains 25 recordings (19 males and 6 females) from healthy adult volunteers aged from 18 to 65 years. The recordings were collected at the French Armed Forces Biomedical Research Institute's (IRBA) Fatigue and Vigilance Unit (Bretigny-Sur-Orge, France). Each recording includes the EEG derivations C3-M2, C4-M1, F3-F4, F3-M2, F3-O1, F4-O2, O1-M2, O2-M1, one EMG channel, left/right EOG channels and one ECG channel. The recordings are sampled at 512 Hz.

**DOD-O.** The dataset contains 55 recordings (35 males and 20 females) from patients suffering from obstructive sleep apnea (OSA) aged from 39 to 62 years. The recordings were collected at the Stanford Sleep Medicine Center. Each recording includes the EEG derivations C3-M2, C4-M1, F4-M1,

F3-F4, F3-M2, F3-O1, F4-O2, FP1-F3, FP1-M2, FP1-O1, FP2-F4, FP2-M1, FP2-O2, one EMG channel, left/right EOG channels and one ECG channel. The recordings are sampled at 250 Hz.

We only consider the single-channel EEG C4-M1 to train our DSN-L architecture, and we use all the available channels to train SSN architecture, on both DOD-H and DOD-O. As in [23], a band-pass Butterworth IIR filter is applied between [0.4, 18] Hz to remove residual PSG noise, and the signals are resampled at 100 Hz. The signals are then clipped and divided by 500 to remove extreme values. The recordings from both DOD-H and DOD-O datasets are scored by five physicians from three different sleep centers according to the AASM rules [2].

DOD-H and DOD-O contain the following annotations $W$, $N1$, $N2$, $N3$, $R$, and $NC$, where $NC$ is a not classified epoch. All the scorers agree about the $NC$ epochs (100% of agreement). Therefore, all of them are removed from the data. Unlike the previous IS-RC database, for each epoch five annotations are always available.

In Table 1 we report a summary of the total number and percentage of the epochs per sleep stage for the DOD-H, DOD-O and IS-RC datasets.

## Deep learning-based scoring architectures

**DSN-L** [32] is a simplified *feed-forward* version of the original DeepSleepNet by [24]. Unlike the original network, in [32] we proposed to employ only the first *representation learning* block, and we proposed to simply train it with a *sequence-to-epoch* learning approach. The architecture receives in input a sequence of 90-second epochs, and it predicts the corresponding target of the central epoch of the sequence, i.e., many-to-one or sequence-to-epoch classification scheme. The *representation learning* architecture consists of two parallel convolutional neural networks (*CNNs)* branches, with small $CNN_{\theta Small}$ and large $CNN_{\theta Large}$ filters at the first layer. The principle is to extract high-time resolution patterns with the small filters, and to extract high-frequency resolution patterns with the large ones. This idea comes from the way the signal processing experts define the trade-off between temporal and frequency precision in the feature extraction procedure [33]. Each CNN branch consists of four convolutional layers and two max-pooling layers. Each convolutional layer executes three basic operations: 1-Dimensional convolution of the filters with the sequential input; batch normalization [34]; element-wise rectified linear unit (ReLU) activation function. Then the pooling layers are used to downsample the input. In Figure 1 we report an overview of the architecture, with details about the filter size, the number of filters and the stride size of each convolutional layer. The pooling size and the stride size for each pooling layer are also specified.

**SSN** [23] consists of two main parts as shown in Figure 2: (i) The *epoch encoder* part, inspired by [22], or what we refer to as epoch processing block (*EPB*), is designed to process 30-second multi-channel EEG epochs, and it aims at learning epoch-wise features. (ii) The *sequence encoder* part, inspired by [24], or what we refer to as sequence processing block (*SPB*), is designed to process sequences of epochs, and it aims to encode the temporal information (e.g., stage transition rules). The *SPB* block consists of two layers of bidirectional gated recurrent unit (GRU) with skip-connections (SkipGRU) and the final classification layer. The architecture receives in input a sequence of PSG epochs, specifically temporal context is set to twenty-one, and it outputs the corresponding sequences of sleep stages at once, i.e., many-to-many or sequence-to-sequence classification scheme.

In both, DSN-L and SSN, the softmax function and the cross-entropy loss function $H$ (see Supplementary Analyses) are used to train the models to output the probabilities $\hat{p}_{i,k}$ for the five mutually exclusive classes $K$, that correspond to the five sleep stages. The cross-entropy loss quantify the agreement between the prediction $\underline{p}_i$ and the target $\underline{y}_i$ (i.e., sleep stage label) for each sleep epoch. The aim is to minimize the cross-entropy loss function $H$, i.e., minimize the distance between the prediction $\underline{p}_i$ and the target $\underline{y}_i$.

The models are trained end-to-end via backpropagation, using mini-batch Adam gradient-based optimizer [35], with a learning rate $lr$. The training procedure runs up to a maximum number of iterations (e.g., 100 iterations), as long as the break early stopping condition is satisfied (i.e., the validation F1-score stopped improving after more than a certain epochs; the model with the best validation F1-score is used at test time). All the training parameters (e.g., adam-optimizer parameters beta1 and beta2, mini-batch size, learning rate etc.) are all set as recommended in [32] and [23].

In Supplementary Analyses we also report additional mathematical details about both the scoring architectures.

## Consensus in multi-scored datasets

Inspired by [23,27], we evaluate the performance of the sleep scoring architectures, as well as the performance of each physician, using the consensus among the five/six different scorers. The majority vote from the scorers has been computed - i.e., we assign to each 30-second epoch the most voted sleep stage among the physicians. In case of ties, we consider the label from the most reliable scorer. The most reliable scorer is the one that is frequently in agreement with all the others. We use the $Soft\text{-}Agreement$ metric proposed in [23] to rank the reliability of each physician, and to finally define the most reliable scorer.

We denote with $J$ the total number of scorers and with $j$ the single-scorer. The one-hot encoded sleep stages given by the scorer $j$ are: $\hat{y}_j \in [0,1]^{K x T}$, where $K$ is the number of classes, i.e., $K = 5$ sleep stages, and $T$ is the total number of epochs. The probabilistic consensus $\hat{z}_j$ among the $J - 1$ scorers ($j$ excluded) is computed using the following:

$$\hat{z}_j = \frac{\sum_{i=1}^{J} \hat{y}_i[t]}{max \sum_{i=1}^{J} \hat{y}_i[t]} \quad \forall t; \quad i \neq j \qquad (1)$$

where $t$ is the $t$-$th$ epoch of $T$ epochs and $\hat{z}_j \in [0,1]^{KxT}$, i.e., 1 is assigned to a stage if it matches the majority or if it is involved in a tie. The $Soft$-$Agreement$ is then computed across all the $T$ epochs as:

$$Soft\text{-}Agreement_j = \frac{1}{T} \sum_{t=0}^{T} \hat{z}_j[y_j] \qquad (2)$$

where $\hat{z}_j[y_j]$ denotes the probabilistic consensus of the sleep stage chosen by the scorer $j$ for the $t$-$th$ epoch. $Soft\text{-}Agreement_j \in [0,1]$, where the zero value is assigned if the scorer $j$ systematically scores all the annotations incorrectly compared to the others, whilst 1 is assigned if the scorer $j$ is always involved in tie cases or in the majority vote. The $Soft$-$Agreement$ is computed for all the scorers, and the values are sorted from the highest - high reliability - to the lowest - low reliability. The $Soft$-$Agreement$ is computed for each patient, i.e., the scorers are ranked for each patient, and in case of a tie the top-1 physician will be the one used for that patient.

## Label smoothing with *soft-consensus*

The predicted sleep stage for each 30-second epoch is associated to a probability value $\hat{p}_i$, which should mirror its ground truth correctness likelihood. When this happens, we can state that the model is well calibrated, or that the model provides a *calibrated confidence* measure along with its prediction [36]. Consider, for example, a model trained to classify images as either containing a dog or not; out of ten test set images it outputs the probability of there being a dog as 0.60 for every image. The model is perfectly calibrated if six dog images are present in the test set. Label smoothing [37] has been shown to be a suitable technique to improve the calibration of the model.

By default, the cross-entropy loss function $H$ is computed between the prediction $\underline{p_i}$ and the target $\underline{y_i}$ (i.e., the one-hot encoded sleep stages, 1 for the correct class and 0 for all the other classes). Whenever a model is trained with the label smoothing technique, the hard target is usually smoothed with the standard *uniform* distribution $1/K$ *(3)*. Thus, the cross-entropy loss function *(4)* is minimized by using the weighted mixture of the target $y_{i,k}^{LS_U}$.

$$y_{i,k}^{LS_U} = y_{i,k} \cdot (1-\alpha) + \alpha \cdot 1/K \qquad (3)$$

$$H(\underline{y_i}, \underline{p_i}) = \sum_{k=1}^{K} -y_{i,k}^{LS_U} \cdot log(\hat{p}_{i,k}) \qquad (4)$$

where $\alpha$ is the smoothing parameter, $K$ the number of sleep stages, $y_{i,k}^{LS_U}$ the weighted mixture of the target and $\hat{p}_{i,k}$ the output of the model with the predicted probability values.

In our study, we exploit the label smoothing technique to improve the insertion of the knowledge from the multiple-scorers in the learning process. We propose to use the *Soft-Consensus (5)* as our new distribution to smooth the hard target $y_{i,k}$.

$$Soft\text{-}Consensus_i = \frac{\#(Y_i = y_{i,k})}{M} \qquad (5)$$

where $Y_i$ is the set of observations - i.e., annotations given by the different physicians - for the $i\text{-}th$ epoch, $k$ is the class index, $M$ is the number of observations and $\#$ is the cardinality of the set $(Y_i = y_{i,k})$. In simple words, the probability value for each sleep stage $k$ is computed as the sum of its occurrences divided by the total number of observations.

$Soft\text{-}consensus_{i,k} \in [0,1]^{1xK}$ is the one-dimensional vector that we use to smooth the hard target *(6)*, and then minimize the cross-entropy loss function *(7)*.

$$y_{i,k}^{LS_{SC}} = y_{i,k} \cdot (1 - \alpha) + \alpha \cdot soft\text{-}consensus_{i,k} \quad (6)$$

$$H(\underline{y_i}, \underline{p_i}) = \sum_{k=1}^{K} - y_{i,k}^{LS_{SC}} \cdot log(\hat{p}_{i,k}) \quad (7)$$

To make it clearer, we report a practical example on how to compute the *soft-consensus* distribution, and how to exploit it to smooth our labels. Consider the following set of observations $Y_i = [W, W, W, N1, N2]$ given by five different physicians for the same $i\text{-}th$ epoch.

We can calculate the $soft - consensus$ consensus as following:

$$Soft\text{-}Consensus_{i,k} = [\,p_W = 3/5\,,\, p_{N1} = 1/5\,,\, p_{N2} = 1/5\,,\, p_{N3} = 0/5\,,\, p_{REM} = 0/5\,]$$

$$Soft\text{-}Consensus_{i,k} = [\,0.6, 0.2, 0.2, 0, 0\,]$$

By applying *(5)* and *(6)* we obtain the following $y_{i,k}^{LS_{SC}}$ smoothed hard-target with $\alpha = 0.5$:

$$y_{i,k}^{LS_{SC}} = y_{i,k} \cdot (1 - \alpha) + \alpha \cdot Soft\text{-}Consensus_{i,k} = [0.8, 0.1, 0.1, 0, 0]$$

that corresponds to the one-hot encoded target:

$$y_{i,k} = [\, p_W = 1,\ p_{N1} = 0,\ p_{N2} = 0,\ p_{N3} = 0,\ p_{REM} = 0]$$

We perform a simple grid-search to set the smoothing hyperparameter $\alpha$. When the model is trained with the labels smoothed by the *uniform* distribution the $\alpha$ value ranges between (0,0.5] with step 0.1. Extreme values are not considered as for $\alpha = 0$ the model is trained using the standard hot-encoding vector; whilst for values higher than 0.5, e.g., $\alpha = 1$, the model would be trained using mainly/only the *uniform* distribution $1/K$ for each sleep stage. When the model is trained with the labels smoothed by the $Soft\text{-}Consensus$ distribution the $\alpha$ value ranges between (0,1] with step 0.1. In the latter case we also investigate an $\alpha$ value equal to 1 to evaluate the full impact of the consensus distribution on the learning procedure.

## Experimental design

We evaluate DSN-L and SSN using the $k$-fold cross-validation scheme. We set $k$ equal to 10 for IS-RC, 25 for DOD-H (leave-one-out evaluation procedure) and 10 for DOD-O datasets, consistent with what was done in [23].
In Table 2 we summarize the data split for each dataset.

The following experiments are conducted on both DSN-L and SSN models for each dataset:

- ***base***. The models are trained without label smoothing.

- **base+LS$_U$**. The models are trained with label smoothing using the standard $1/K$ uniform distribution - i.e., the hard targets (scorer consensus) are weighted with the *uniform* distribution.

- **base+LS$_{SC}$**. The models are trained with label smoothing using the proposed *soft-consensus* - i.e., the hard targets (scorer consensus) are weighted with the *soft-consensus* distribution.

These models, differently trained, have been evaluated with and without *MC* dropout ensemble technique. In Table 4, Table 5 and Table 6 section **Results** we present the results obtained for each experiment on both DSN-L and SSN evaluated on IS-RC, DOD-H and DOD-O datasets.

## Metrics

### *Performance.*

The per-class F1-score, the overall accuracy (Acc.), the macro-averaging F1-score, the weighted-averaging F1-score (i.e., the metric is weighted by the number of true instances for each label, so as to consider the high imbalance between the sleep stages) and the Cohen's kappa have been computed per-subject from the predicted sleep stages from all the folds to evaluate the performance of our model [38, 39].

### *Hypnodensity graph.*

The hypnodensity-graph is an efficient visualization tool introduced in [27] to plot the probability distribution over each sleep stage for each 30-second epoch over the whole night. Unlike the standard hypnogram sleep cycle visualization tool, the hypnodensity-graph shows the probability of occurrence of each sleep stage for each 30-second epoch; so it is not limited to the discrete sleep stage value (see Figure 3).

In our study we have used the hypnodensity-graph to display both the model output - i.e., the probability vectors $\hat{p}_{i,k}$ - and the multi-scorer $Soft\text{-}Consensus_{i,k}$ probability distributions.

The Averaged Cosine Similarity ($ACS$) is used to quantify the similarity between the hypnodensity-graph generated by the model and the hypnodensity-graph generated by the $Soft\text{-}Consensus$. The $ACS$ has been computed as follows:

$$ACS = \frac{1}{N}\sum_{i=1}^{N} \frac{soft\text{-}consensus_{i,k} \cdot \hat{p}_{i,k}}{||soft\text{-}consensus_{i,k}|| \cdot ||\hat{p}_{i,k}||} \quad (8)$$

where $N$ is the number of epochs in the whole night, $||.||$ is the norm computed for the predicted probability vector $\hat{p}_{i,k}$ and the $Soft\text{-}Consensus_{i,k}$ ground-truth vector for the $i\text{-}th$ epoch. Thus, the cosine-similarity is averaged across all the epochs $N$ to obtain our averaged $ACS$ unique score of similarity. The cosine-similarity values may range between 0 i.e., high dissimilarity and 1 i.e., high similarity between the vectors.

### *Calibration.*

The calibration of the model is evaluated by using the expected calibration error ($ECE$) metric proposed in [40]. By ($ECE$) we compute the difference in expectation between the accuracy $acc$ and the $conf$ (i.e., the *softmax* output probabilities) values. More in detail, the predictions are divided into $M$ equally spaced bins (with size $1/M$), then we compute the accuracy $acc(B_m)$ and the average predicted probability value $conf(B_m)$ for each bin as follows:

$$acc(B_m) = \frac{1}{|B_m|} \cdot \sum_{i \in B_m} 1(\hat{y}_i = y_i) \quad (9)$$

$$conf(B_m) = \frac{1}{|B_m|} \cdot \sum_{i \in B_m} \hat{p}_i \quad (10)$$

where $y_i$ is the true label and $\hat{y}_i = argmax(\hat{p}_{i,k})$ is the predicted label for the $i$-$th$ epoch; $B_m$ is the group of samples whose predicted probability values fall in $I_m = (\frac{m-1}{M}, \frac{m}{M}]$ and $\hat{p}_i = max(\hat{p}_{i,k})$ is the predicted probability value for sample the $i$-$th$ 30-second epoch. Finally, the $ECE$ value is computed as the weighted average of the difference between the $acc$ and the $conf$ among the $M$ bins:

$$ECE = \sum_{m=1}^{M} \frac{|B_m|}{n_B} \cdot |acc(B_m) - conf(B_m)| \quad (11)$$

where $n_B$ is the number of samples in each bin. Perfectly calibrated models have $acc(B_m) = conf(B_m)$ for all $m \in \{1, \ldots, M\}$, resulting in $ECE = 0$.

## Results

In Table 3 we first report for all the multi-scored databases IS-RC, DOD-H and DOD-O, the overall scorers performance and their $Soft - Agreement$ ($SA$), i.e., the agreement of each scorer with the consensus among the physicians. On IS-RC we have on average a lower inter-scorer agreement ($SA$ equal to 0.69, with an F1-score 69.7%) compared to both DOD-H and DOD-O ($SA$ equal to 0.89 and 0.88, with an F1-score 88.1% and 86.4% respectively). Consequently, we expect a higher efficiency of our label smoothing with the *soft-consensus* approach (*base+LS<sub>SC</sub>*) on the experiments conducted on the IS-RC database. The lower the inter-scorer agreement, the lower should be the performance of a model trained with the one-hot encoded labels (i.e., the majority vote weighted by the degree of consensus from each physician).

In Table 4 and Table 5 we report the overall performance, the calibration measure and the hypnodensity similarity measure of the three different DSN-L and SSN models on the three databases IS-RC, DOD-H and DOD-O. The performance of the DSN-L *base* models are higher compared to the performance averaged among the scorers on the IS-RC database, but not on the DOD-H and DOD-O databases. In contrast, the performance of the SSN *base* models are always higher than the performance averaged among the scorers on all the databases. We highlight that the results we report for SSN on DOD-H and DOD-O are slightly different compared to the one reported in [23]. We decided to not compute a weight (from 0 to 1) for each epoch, based on how many scorers voted for the consensus. We do not balance the importance of each epoch when we compute the above mentioned metrics. We think it is unfair to constrain any metrics based on the amount of voting physicians. Overall, the results show an improvement in performance on all the databases (i.e overall accuracy, MF1-score, Cohen's kappa ($k$), and F1-score) from the baseline (*base*) and the label smoothing with the *uniform* distribution (*base+LS$_U$*) models, to the ones trained with label smoothing along with the proposed *soft-consensus* distribution (ie. *base+LS$_{SC}$*).

The $ACS$ is the metric that best quantifies the ability of the model in adapting to the consensus of the group of scorers. A higher $ACS$ value means a higher similarity between the hypnodensity-graph generated by the model and the hypnodensity-graph generated by the *soft-consensus* (i.e., the model better adapts to the consensus of the group of physicians). As all the other metrics the $ACS$ value is computed per subject, but here we report the mean and also the standard deviation across subjects ($\mu \pm \sigma$). We found a significant improvement in the $ACS$ value from the *base* and the *base+LS$_U$* models to the *base+LS$_{SC}$* models on all the databases and on both DSN-L (p-values < 0.01) and SSN (p-values < 0.05). Hence, our approach enables both DSN-L and SSN architectures to significantly adapt to the group consensus on all the multi-scored datasets.

We could easily infer that the SSN architecture is better (i.e., higher performance) compared to our DSN-L architecture. The purpose of our study is not to highlight whether one architecture is better than the other, but we can not fail to notice the high values of confidence (the $conf$ value is the average of the softmax output max-probabilities) obtained on the SSN based models. High values of confidence still persist despite smoothing the labels (with both *uniform* and soft-consensus distributions) during the training procedure. The SSN architecture is not highly responsive to the changes in probability values we implemented on the one-hot encoded labels. It always rely/overfit on the $max$ probability value given for each epoch, i.e., the consensus among the five/six different scorers. Indeed, on the IS-RC, which is the database with the lower inter-scorer agreement, the SSN *base+LS$_{SC}$* model reaches a higher value of F1-score, i.e., 81.6%, compared to our DSN-L *base+LS$_{SC}$* model, i.e., 75.9% , but a lower value of $ACS$ (0.817 on SSN and 0.836 on DSN-L, with a p-value < 0.01). The SSN model overfit to the majority vote or the $max$ probability value given for each epoch, whilst the DSN-L better adapts to the consensus of the group of scorers (i.e., better encodes the variability among the physicians).

The last statement is also strengthened by the Supplementary Figure S1 and Figure S2. For DSN-L and SSN we report the $ACS$ values across all the experimented $\alpha$ values, on both the *base+LS$_U$* and the *base+LS$_{SC}$* models tested on the three databases. As expected, the DSN-L model shows a high sensitivity in $ACS$ values to changes in α-hyperparameter across all databases. This sensitivity is not as strong with the SSN model.

Moreover, we want to stress that the standard *uniform* distribution is not as efficient as the proposed *soft-consensus* distribution in encoding the scorer's variability. By using the *uniform* distribution we are not able to learn as well the complexity of the degree of agreement between the different physicians. Indeed, in Supplementary Figure S1, on the DSN-L model, we clearly show how the $ACS$ value proportionally increases with the α-hyperparameter only by using the proposed *soft-* *consensus* distribution. In Figure 4 we also show, on a patient from the DOD-O dataset, how we achieve a higher $ACS$ value with the proposed *base+LS$_{SC}$* model with the *soft-consensus* distribution, compared to *base+LS$_U$* model with the standard *uniform* distribution. The graph clearly highlights the differences between the output probabilities predicted by the different models. The probabilities predicted using our approach *base+LS$_{SC}$* (d) are closer to the ground-truth (a) compared to the ones predicted from the other models (e.g. refer to min. 300 and to the probabilities associated with the sleep stage N3).

## Discussion

Many deep learning based approaches are available and from a technical point of view there is not that much that is left to be done to improve their performance. It is not reasonable to reach a performance higher than the gold standard that is used to train the architectures. Infact, the real limitation is the low inter-rater agreement due to subjective interpretation.

Therefore in this paper we focus on how to better integrate the inter-rater agreement information into the automated sleep scoring algorithms. Presently, information about the variability is not completely exploited. The algorithms are trained on the majority vote consensus, leading to overfitting on the majority vote weighted by the degree of consensus from each physician.

We introduce a more complete methodology to integrate scorer's variability in the training procedure. We demonstrate the efficiency of label smoothing along with the *soft-consensus* distribution in encoding the scorers's variability into the training procedure of both DSN-L and SSN scoring algorithms. The results show an improvement in overall performance from the *base* models to the ones trained with *base+LS$_{SC}$*. We introduce the averaged cosine similarity metric to better quantify the similarity between the probability distribution predicted by the models and the ones generated by the scorer consensus. We obtain a significant improvement in the $ACS$ values from the *base* models to the *base+LS$_{SC}$* models on both DSN-L and SSN architectures. Based on the reported high confidence values, we found that SSN tends to overfit on each dataset. Specifically, it tends to overfit on the majority vote weighted by the degree of consensus from each physician, but does not encode as well their variability.

To our knowledge, our work is the first attempt to transfer the variability, the uncertainty and the noise among multiple-scorers to an automated sleep scoring system.

We have proved the strength of our approach and especially the use of the soft-consensus distribution by comparing it with the *base* models and the implemented models trained with label smoothing but using the uniform distribution. We clearly show on all the experiments the higher overall performance and $ACS$ values achieved with the soft-consensus distribution.

In order to generalize our approach, there are two big limitations. The first is that a far bigger datasets, highly heterogeneous (with different diagnosis, age range, gender etc.) scored by multiple scorers would be necessary. The second is that the recordings exploited in this study are not labeled by a homogeneous group of board certified sleep scorers. Further studies should be carried out to better quantify the resilience and the reproducibility of the proposed approach. To achieve a high-performance sleep scoring algorithm, we must take into account both the variability of the recordings and the variability between the different sleep scorers. We should train our sleep scoring models on PSG recordings from different large-scale-heterogeneous data cohorts, and ideally with each recording scored by multiple physicians.

In summary, the possibility of exploiting the full set of information that is hidden in a multi-scored dataset would certainly enhance automated deep learning algorithms performance. The present approach enables us to better adapt to the consensus of the group of scorers, and, as a consequence, to better quantify the disagreement we have between the different scorers. The proposed approach results quite effective in encoding the complexity of the scorers' consensus within the classification algorithm, whose importance is often underestimated.

## Funding

Prof. F. D. Faraci was supported by SPAS: Sleep Physician Assistant System project, from Eurostars funding programme. Prof. P. Favaro was supported by the IRC Decoding Sleep: From Neurons to Health and Mind, from the University of Bern, Switzerland.

## Disclosure Statement

Conflicts of interest. The authors declare that they have no conflict of interest.

Financial Disclosure: none. Non financial Disclosure: none.

# Figure Captions

**Figure 1.** DeepSleepNet-Lite architecture.

An overview of the *representation learning* architecture from [24], with our *sequence-to-epoch* training approach.

**Figure 2.** SimpleSleepNet architecture.

An overview of the SimpleSleepNet architecture from [23]. $h_{t-1}, h'_{t-1}$ represent the hidden states of the GRU layers from the previous epoch of the sequence and $h_{t+1}, h'_{t+1}$ the hidden states of the GRU layers from the next epoch of the sequence. $a_t$ is the embedding of the current epoch.

**Figure 3.** Hypnogram and hypnodensity-graph from the scorers labels.

Example of hypnogram and hypnodensity-graph for a subject from the DOD-H with the highest percentage 14% of N1 sleep stages. For each 30-second epoch we report on top the hypnogram, i.e., the discrete sleep stage values (majority vote from the scorers labels); on bottom the hypnodensity-graph, i.e., the cumulative probabilities of each sleep stage ( soft- consensus computed from the scorers labels). The hypnodensity-graph allows us to better appreciate the low level of agreement of a specific sleep stage among the different scorers. In this example, the sleep stages N1 are often associated with a high percentage of residual probability in awake or N2, thus at the transitions from one sleep stage to another.

**Figure 4.** Hypnodensity-graphs from the scorers labels and from the predicted probabilities from the experimented models.

Example of hypnodensity-graphs for a subject from the DOD-O. (a) Soft-consensus computed from the scorers labels; (b) DSN-L *base* model; (c) DSN-L *base+LS$_U$*; (d) DSN-L *base+LS$_{SC}$*. We also report the ACS value computed between the hypnodensity-graph associated to soft-consensus and the ones generated from the predicted probabilities of each model. We reach a higher $ACS$ value with the proposed *base+LS$_{SC}$* model with the *soft-consensus* distribution (d), compared to the baseline (b) and the *base+LS$_U$* model with the standard *uniform* distribution (c).

# Tables

**Table 1**

Number and percentage of 30-second epochs per sleep stage for the IS-RC, DOD-H and DOD-O datasets.

|  | W | N1 | N2 | N3 | R | Total |
|---|---|---|---|---|---|---|
| **IS-RC** | 24517 (29.1%) | 3773 (4.5%) | 40867 (48.5%) | 3699 (4.4%) | 11475 (13.6%) | 84331 |
| **DOD-H** | 3075 (12.5%) | 1463 (5.9%) | 12000 (48.7%) | 3442 (14.0%) | 4685 (19.0%) | 24665 |
| **DOD-O** | 10520 (19.8%) | 2739 (5.1%) | 26213 (49.2%) | 5617 (10.6%) | 8147 (15.3%) | 53236 |

**Table 2**

Data split on the IS-RC, DOD-H and DOD-O datasets.

| | Size | Experimental Setup | Held-out Validation Set | Held-out Test Set |
|---|---|---|---|---|
| **IS-RC** | 70 | 10-fold CV | 13 subjects | 7 subject |
| **DOD-H** | 25 | 25-fold CV | 6 subjects | 1 subjects |
| **DOD-O** | 55 | 10-fold CV | 12 subjects | 6 subjects |

**Table 3**

Scorers performance on IS-RC, DOD-H and DOD-O datasets with $Soft-Agreement$ ($SA$), overall accuracy (%Acc.), macro F1-score (%MF1), Cohen's Kappa ($k$), weighted-averaging F1-score (%F1) and % per-class F1-score. The scorer with the best performance (i.e., high agreement with the consensus among the different physicians) is indicated in bold.

| | | | Overall Metrics | | | | Per-Class F1-Score | | | | |
|---|---|---|---|---|---|---|---|---|---|---|---|
| | Scorers | $SA$ | Acc. | MF1 | $k$ | F1 | W | N1 | N2 | N3 | R |
| **IS-RC** | Scorer-1 | 0.79 | 83.0 | 69.5 | 0.72 | 83.8 | 83.1 | 47.2 | 87.3 | 48.0 | 82.1 |
| | **Scorer-2** | 0.81 | 89.4 | 72.8 | 0.82 | 89.2 | 91.3 | 57.6 | 92.5 | 32.9 | 89.8 |
| | Scorer-3 | 0.53 | 40.7 | 26.5 | 0.11 | 40.8 | 29.8 | 14.7 | 54.5 | 17.9 | 15.6 |
| | Scorer-4 | 0.52 | 38.9 | 26.1 | 0.12 | 40.5 | 28.6 | 14.7 | 54.2 | 15.4 | 17.5 |
| | Scorer-5 | 0.70 | 73.7 | 61.6 | 0.63 | 75.8 | 88.7 | 36.9 | 70.2 | 25.8 | 86.2 |
| | Scorer-6 | 0.79 | 87.2 | 77.2 | 0.81 | 88.2 | 92.5 | 54.6 | 89.4 | 59.8 | 89.5 |
| | Average | 0.69 | 68.7 | 55.5 | 0.53 | 69.7 | 68.9 | 37.6 | 74.7 | 33.3 | 63.5 |
| **DOD-H** | Scorer-1 | 0.88 | 87.0 | 81.5 | 0.81 | 87.4 | 87.5 | 60.0 | 89.4 | 84.8 | 85.7 |
| | Scorer-2 | 0.91 | 89.3 | 84.1 | 0.84 | 89.7 | 87.4 | 65.1 | 91.6 | 84.3 | 92.2 |
| | **Scorer-3** | 0.92 | 90.6 | 84.5 | 0.86 | 90.4 | 89.9 | 67.5 | 92.1 | 77.9 | 95.3 |
| | Scorer-4 | 0.84 | 82.6 | 76.7 | 0.75 | 83.1 | 76.5 | 49.1 | 85.4 | 80.7 | 92.0 |
| | Scorer-5 | 0.92 | 89.9 | 83.6 | 0.85 | 89.9 | 86.7 | 66.0 | 92.1 | 81.0 | 92.2 |
| | Average | 0.89 | 87.9 | 82.1 | 0.82 | 88.1 | 85.5 | 61.5 | 90.0 | 81.7 | 91.5 |
| **DOD-O** | Scorer-1 | 0.87 | 85.0 | 75.1 | 0.77 | 84.6 | 90.0 | 49.5 | 85.2 | 67.6 | 83.3 |

| | | | | | | | | | | |
|---|---|---|---|---|---|---|---|---|---|---|
| | Scorer-2 | 0.87 | 85.0 | 78.2 | 0.78 | 86.0 | 89.3 | 58.4 | 85.4 | 69.1 | 88.6 |
| | Scorer-3 | 0.88 | 86.0 | 75.0 | 0.78 | 84.6 | 91.0 | 54.3 | 86.5 | 56.1 | 87.0 |
| | Scorer-4 | 0.88 | 86.7 | 77.7 | 0.80 | 87.2 | 91.2 | 59.3 | 89.4 | 62.9 | 85.8 |
| | **Scorer-5** | 0.91 | 89.9 | 82.3 | 0.84 | 90.0 | 93.7 | 68.3 | 90.7 | 70.5 | 88.2 |
| | Average | 0.88 | 86.5 | 77.6 | 0.79 | 86.4 | 91.0 | 58.0 | 87.3 | 65.2 | 86.5 |

**Table 4**

Overall metrics, per-class F1-score, calibration and $ACS$ hypnodensity graph similarity measures of the DSN-L models obtained from 10-fold cross-validation on IS-RC dataset, from 25-fold cross-validation on DOD-H dataset, and from 10-fold cross-validation on DOD-O dataset. Best shown in bold.

| | Models | $\alpha$ | Overall Metrics | | | | Per-Class F1-Score | | | | | Calibration | | Hypn. |
|---|---|---|---|---|---|---|---|---|---|---|---|---|---|---|
| | | | Acc. | MF1 | $k$ | F1 | W | N1 | N2 | N3 | R | ECE | $conf.$ | $ACS$ |
| IS-RC | base | - | 69.6 | 50.6 | 0.56 | 70.0 | 81.6 | 11.8 | 71.9 | 27.2 | 60.7 | **0.096** | 79.0 | 0.772 ± 0.075 |
| | base+$LS_U$ | 0.4 | 74.8 | **57.0** | 0.63 | 75.8 | 83.3 | **24.3** | 79.0 | 30.6 | **67.7** | 0.296 | 45.2 | 0.806 ± 0.042 |
| | base+$LS_{SC}$ | 0.6 | **75.8** | 56.5 | **0.69** | 75.9 | 83.5 | 19.5 | **79.7** | 33.3 | 66.4 | 0.190 | 56.7 | **0.836 ± 0.041** |
| DOD-H | base | - | 76.9 | 70.0 | 0.68 | 77.2 | 79.7 | 39.5 | 78.8 | 76.5 | 75.2 | 0.163 | 92.7 | 0.817 ± 0.097 |
| | base+$LS_U$ | 0.2 | 75.3 | 68.7 | 0.66 | 75.2 | 78.8 | 40.0 | 75.9 | 72.0 | 76.8 | 0.059 | 68.9 | 0.829 ± 0.068 |
| | base+$LS_{SC}$ | 0.8 | **80.2** | **72.4** | **0.72** | **80.4** | **80.4** | **42.3** | **83.4** | 77.6 | **78.8** | **0.016** | 81.4 | **0.873 ± 0.053** |
| DOD-O | base | - | 77.3 | 67.8 | 0.66 | 78.0 | 80.7 | 41.2 | 81.0 | 68.1 | 68.3 | 0.131 | 90.2 | 0.840 ± 0.073 |
| | base+$LS_U$ | 0.1 | 77.5 | 68.0 | 0.67 | 78.2 | **80.8** | 41.9 | 80.4 | 68.4 | **68.7** | 0.009 | 78.4 | 0.859 ± 0.072 |
| | base+$LS_{SC}$ | 1 | **79.4** | **69.6** | **0.69** | **79.9** | 80.4 | **43.8** | **83.5** | **72.5** | 68.1 | **0.009** | 78.3 | **0.878 ± 0.061** |

**Table 5**

Overall metrics, per-class F1-score, calibration and $ACS$ hypnodensity graph similarity measures of the SSN models obtained from 10-fold cross-validation on IS-RC dataset, from 25-fold cross-validation on DOD-H dataset, and from 10-fold cross-validation on DOD-O dataset. Best shown in bold.

| | Models | $\alpha$ | Overall Metrics | | | | Per-Class F1-Score | | | | | Calibration | | Hypn. |
|---|---|---|---|---|---|---|---|---|---|---|---|---|---|---|
| | | | Acc. | MF1 | $k$ | F1 | W | N1 | N2 | N3 | R | ECE | $conf.$ | $ACS$ |
| IS-RC | $base$ | - | 81.8 | **60.8** | 0.72 | 80.8 | 86.3 | **29.9** | 85.3 | **24.3** | 78.1 | 0.174 | 99.4 | 0.806 ± 0.052 |
| | $base+LS_U$ | 0.3 | 82.5 | 59.8 | 0.72 | 81.1 | 86.5 | 28.8 | 86.5 | 18.7 | 78.7 | 0.169 | 99.3 | 0.811 ± 0.058 |
| | $base+LS_{SC}$ | 0.7 | **83.1** | 60.2 | **0.73** | **81.6** | **86.7** | 27.6 | **86.8** | 20.1 | **79.8** | **0.162** | 99.2 | **0.817 ± 0.047** |
| DOD-H | $base$ | - | 87.1 | 80.2 | 0.81 | 87.1 | 83.6 | 55.5 | 90.0 | **83.3** | 89.0 | 0.126 | 99.7 | 0.890 ± 0.047 |
| | $base+LS_U$ | 0.4 | 87.6 | 81.0 | 0.81 | 87.5 | 85.5 | 57.3 | 90.2 | 82.1 | 90.3 | 0.120 | 99.5 | 0.899 ± 0.034 |
| | $base+LS_{SC}$ | 0.5 | **88.8** | **82.3** | **0.83** | **88.7** | **86.4** | 58.8 | **90.9** | 83.2 | **92.1** | **0.108** | 99.6 | **0.907 ± 0.039** |
| DOD-O | $base$ | - | 85.3 | 75.9 | 0.77 | 85.2 | 88.2 | 50.4 | 87.1 | 65.9 | 88.0 | 0.145 | 99.7 | 0.889 ± 0.056 |
| | $base+LS_U$ | 0.1 | 85.6 | 75.8 | 0.78 | 85.2 | 88.2 | **51.2** | 87.3 | 64.3 | 88.4 | 0.141 | 99.6 | 0.893 ± 0.052 |
| | $base+LS_{SC}$ | 1 | **86.8** | **77.7** | **0.79** | **86.7** | **89.0** | 51.0 | **88.3** | 69.3 | **91.1** | **0.125** | 99.2 | **0.906 ± 0.043** |

**Table 6**

Overall metrics and $ACS$ hypnodensity graph similarity measures on the DSN-L and SSN $base+LS_{SC}$ models, obtained from 10-fold cross-validation on IS-RC dataset, from 25-fold cross-validation on DOD-H dataset, and from 10-fold cross-validation on DOD-O dataset with and without $MC$. Best shown in bold.

|  |  |  | Overall Metrics | | | | Hypn. |
|---|---|---|---|---|---|---|---|
|  |  |  | Acc. | MF1 | $k$ | F1 | $ACS$ |
| **IS-RC** | DSN-L | w/o MC | 75.8 | 56.5 | 0.69 | 75.9 | 0.836 ± 0.041 |
|  |  | w/ $MC$ | **78.6** | **57.6** | **0.67** | **78.0** | **0.850 ± 0.036** |
|  | SSN | w/o MC | **83.1** | **60.2** | 0.73 | **81.6** | 0.817 ± 0.047 |
|  |  | w/ $MC$ | 83.0 | 59.2 | **0.73** | 81.1 | **0.818 ± 0.048** |
| **DOD-H** | DSN-L | w/o MC | 80.2 | 72.4 | 0.72 | 80.4 | 0.873 ± 0.053 |
|  |  | w/ $MC$ | **84.4** | **75.9** | **0.76** | **84.2** | **0.906 ± 0.026** |
|  | SSN | w/o MC | 88.8 | 82.3 | 0.83 | 88.7 | 0.907 ± 0.039 |
|  |  | w/ $MC$ | **89.1** | **82.6** | **0.84** | **89.0** | **0.910 ± 0.039** |
| **DOD-O** | DSN-L | w/o MC | 79.4 | 69.6 | 0.69 | 79.9 | 0.878 ± 0.061 |
|  |  | w/ $MC$ | **80.7** | **70.8** | **0.71** | **80.9** | **0.889 ± 0.059** |
|  | SSN | w/o MC | 86.8 | 77.7 | 0.79 | 86.7 | 0.906 ± 0.043 |
|  |  | w/ $MC$ | **87.1** | **78.0** | **0.80** | **86.9** | **0.909 ± 0.041** |

Figure 1

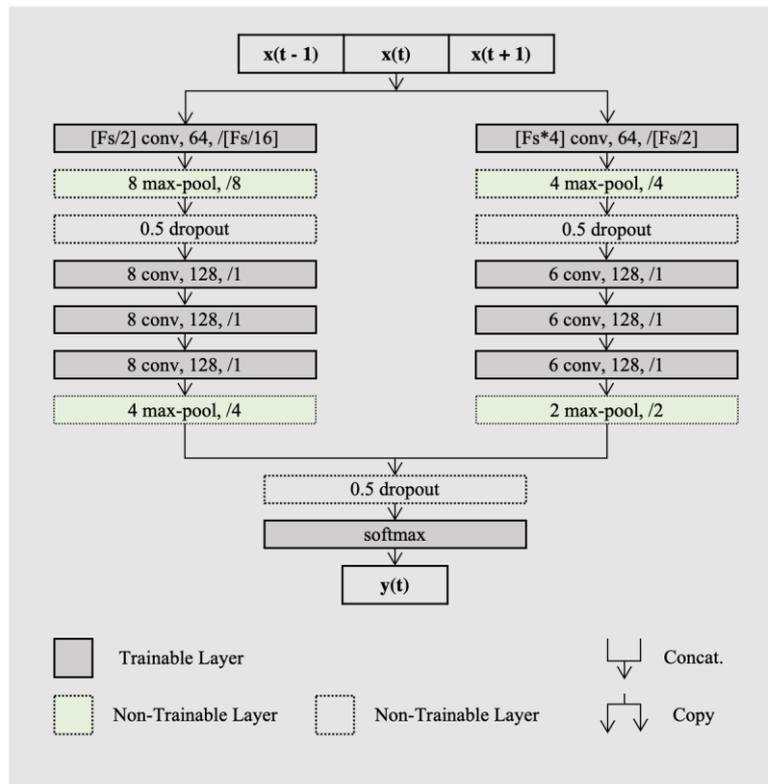

Figure 2

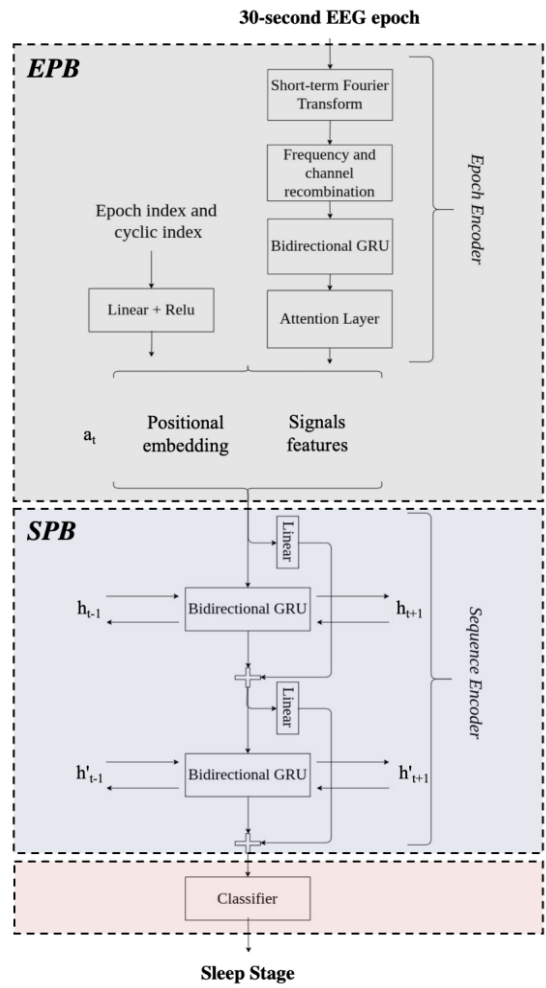

Figure 3

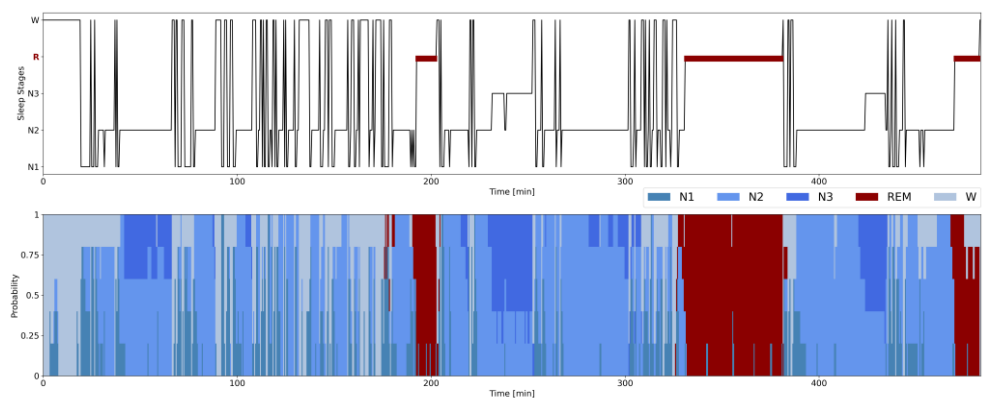

Figure 4

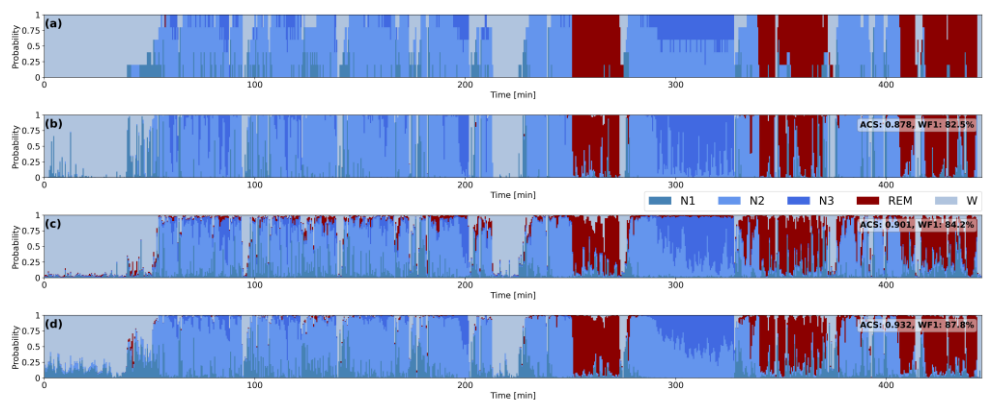